\pdfoutput=1

\documentclass[11pt]{article}

\usepackage{acl}

\usepackage{times}
\usepackage{latexsym}

\usepackage{amsmath}
\usepackage{amssymb}
\usepackage{booktabs}
\usepackage{array,multirow,graphicx}
\usepackage{comment}
\usepackage{hyperref}
\usepackage{xurl}
\usepackage{algorithm}
\usepackage{algorithmic}
\usepackage{graphicx}
\usepackage{xcolor}
\usepackage{xspace}
\usepackage{enumitem}

\usepackage{subcaption}

\usepackage{pifont}
\usepackage{color, colortbl}

\newcommand{\comet}{\textsc{COMET-QE}}

\definecolor{oracle}{HTML}{a61a12}
\definecolor{internal}{HTML}{2c8c11}
\definecolor{external}{HTML}{0b3e85}

\usepackage[T1]{fontenc}

\usepackage[utf8]{inputenc}

\usepackage{microtype}

\usepackage{todonotes}

\title{Detecting and Mitigating Hallucinations in Machine Translation:\\
Model Internal Workings Alone Do Well, Sentence Similarity Even Better}

\author{David Dale \quad \ Elena Voita \quad \ Lo{\"i}c Barrault \quad \ Marta R. Costa-juss\`a  \\
 Meta AI\\ 
  \texttt{\{daviddale,lenavoita,loicbarrault,costajussa\}@meta.com} \\}

\begin{document}
\maketitle
\begin{abstract}



While the problem of hallucinations in neural machine translation has long been recognized, so far the progress on its alleviation is very little. Indeed, recently it turned out that without artificially encouraging models to hallucinate, previously existing methods fall short and even the standard sequence log-probability is more informative. 
It means that characteristics internal to the model can give much more information than we expect, and before using external models and measures, we first need to ask: how far can we go if we \textit{use nothing but the translation model itself}? We propose to use a method that evaluates the percentage of the source contribution to a generated translation. Intuitively, hallucinations are translations ``detached'' from the source, hence they can be identified by low source contribution. This method improves detection accuracy for the most severe hallucinations by a factor of 2 and is able to alleviate hallucinations at test time on par with the previous best approach that relies on external models. Next, if we move away from internal model characteristics and allow external tools, we show that using sentence similarity from cross-lingual embeddings further improves these results. We release an efficient implementation for the considered methods.\footnote{\scriptsize \url{https://github.com/facebookresearch/stopes}}

\end{abstract}

\section{Introduction}

Hallucinations in machine translation (MT) are cases when the model generates output that is partially or fully unrelated to the source sentence.  While generally this phenomenon is not that frequent and has a relatively low impact on corpus-level automatic metrics, the impact of hallucinations on user experience can be rather dramatic. For example, if a machine translation system generates \textit{The staff were very friendly and helpful} in response to an input sentence about e.g. \textit{a marvelous view from the window}, in future a user is unlikely to trust this system.

While the problem of hallucinations is known and important, addressing it is very challenging. Firstly, hallucinations are very rare. This is why previous work mostly resorted to settings where models are encouraged to hallucinate, e.g. artificially perturbing source sentence~\cite{lee2019hallucinations,raunak-etal-2021-curious}, adding specific types of noise to the training data~\cite{raunak-etal-2021-curious}, working under domain shift~\cite{wang-sennrich-2020-exposure,muller-etal-2020-domain}, among others~\cite{zhou-etal-2021-detecting}. Secondly, hallucinations are hard to identify with automatic metrics. For the most part, hallucinations were defined as translations with low quality according to some metric such as e.g. adjusted BLEU or chrF~\cite{lee2019hallucinations,raunak-etal-2021-curious,muller-sennrich-2021-understanding} or translations satisfying some heuristic condition~\cite{berard-etal-2019-naver,raunak-etal-2021-curious}. Overall, it was not clear whether proposed methods indeed detect hallucinations and, if so, whether they transfer to more natural settings. 

Recently, when revisiting previous work in a relatively clean setting, \citet{guerreiro_hallucinations} found that existing methods fall short and the standard sequence log-probability is the most informative. To show this, the authors gathered a large dataset with professional annotations of translations that, according to 10 previously proposed methods, are likely to be hallucinations.
This data (hallucinations along with the model that generated them) made it possible to first, evaluate the performance of various detection methods and second, to work on alleviating hallucinations at test time. For the latter, the idea is ``detect-then-rewrite'': after flagging a translation as likely to be pathological, generate several alternative hypotheses and pick the best one relying on some measure. So far, the best realization of this general framework uses sequence log-probability~-- Seq-Logprob~-- for detection, Monte Carlo dropout~\cite{gal-mcdropout} to generate several alternative translation hypotheses, and \comet{} to pick the final candidate (see \citet{guerreiro_hallucinations} for more details). In this work, we use the same test bed and substantially improve the previous results.

Regarding hallucination detection, we view the observation that Seq-Logprob outperforms previous (specifically targeted to hallucinations) methods as follows: \textit{internal model characteristics may contain much more information than we expect}. Therefore, before developing or using external models and measures, we ask: \textit{how far can we go if we use nothing but the translation model itself}? We propose to use a method that evaluates the percentage of the source contribution to a generated translation. Intuitively, since hallucinations are translations that are ``detached'' from the source (by definition), low source contribution should be able to identify hallucinations. 
Despite the fact that understanding hallucinations was one of the motivations behind the first method evaluating relative source and target contributions~\cite{voita-etal-2021-analyzing}, both
existing methods only looked at highly artificial hallucinations~\cite{voita-etal-2021-analyzing,ferrandoALTI_plus}. We propose to use ALTI+ by~\citet{ferrandoALTI_plus}, the method that aggregates layer-wise tokens attributions, for both hallucination detection and reranking in the ``detect-then-rewrite'' framework. 
For detection of the most severe hallucinations, it is twice more accurate than sequence log-probability. For reranking, it performs on par with the previous best \comet{}. All in all, we show that we can improve the overall pipeline results by solely relying on internal model characteristics.


When allowing external tools, previous work mostly focused on different ways to automatically evaluate quality of a translation example, either with string-based methods or neural quality estimation systems.
This idea (the better we estimate translation quality, the better we are at detecting hallucinations) is natural: hallucinations are low-quality translations in the first place. However, implementing this idea in practice is challenging: even state-of-the-art quality estimation system substantially fails~\cite{guerreiro_hallucinations}. 
We hypothesize that instead of targeting quality evaluation, it might be beneficial to use models trained with a rather different objective. 
Indeed, as we show, similarity between the source and a translation estimated via cross-lingual sentence embeddings outperforms the best internal method. Apart from cross-lingual sentence similarity (which is expected to be sensitive to highly incorrect translations), we find that cross-lingual natural language inference models (less anticipated in the context of machine translation) also perform quite well. To the best of our knowledge, we are the first to apply these models for hallucination detection.

Overall, we show that:
\begin{itemize}
    \item by using only the model's inner workings, we
        \begin{itemize}
        \item[$\circ$] detect the most severe type of hallucinations with twice better precision;
        \item[$\circ$] overwrite hallucinations at test time with results on par with the best previous method that relies on an external model;
        \end{itemize}
    \item models focused on semantic similarity of sentences can detect all types of hallucinations with $80\%$ better precision than the previous methods.

\end{itemize}

\section{Background and Setting}

In this section, we describe the framework and data we use for evaluation of hallucination detection and mitigation methods. This framework was proposed by~\citet{guerreiro_hallucinations} and consists of a large dataset of annotated translations along with the model that produced them. To the best of our knowledge, this is the only released data that can be used to analyze hallucinations in a ``clean'' setting.

\subsection{Model}
\label{sect:model_guerreiro}

The model is Transformer base~\citep{transformer_vaswani} from \texttt{fairseq}~\cite{ott-etal-2019-fairseq} with the standard hyperparameters setting. It was trained on the WMT’18 German-English news translation data excluding Paracrawl~\citep{bojar-etal-2018-findings} -- totalling 5{.}8M sentence pairs. Since \citet{guerreiro_hallucinations} used randomly chosen 1/3 of the dataset as a held-out set for analysis, the model was trained on the remaining 2/3 of the dataset. We use the model released by~\citet{guerreiro_hallucinations}: this is the model that generated hallucinations we analyze.


\subsection{Hallucination Dataset}
\label{sect:dataset}

\begin{figure}[t]
\centering
{\includegraphics[scale=0.28]{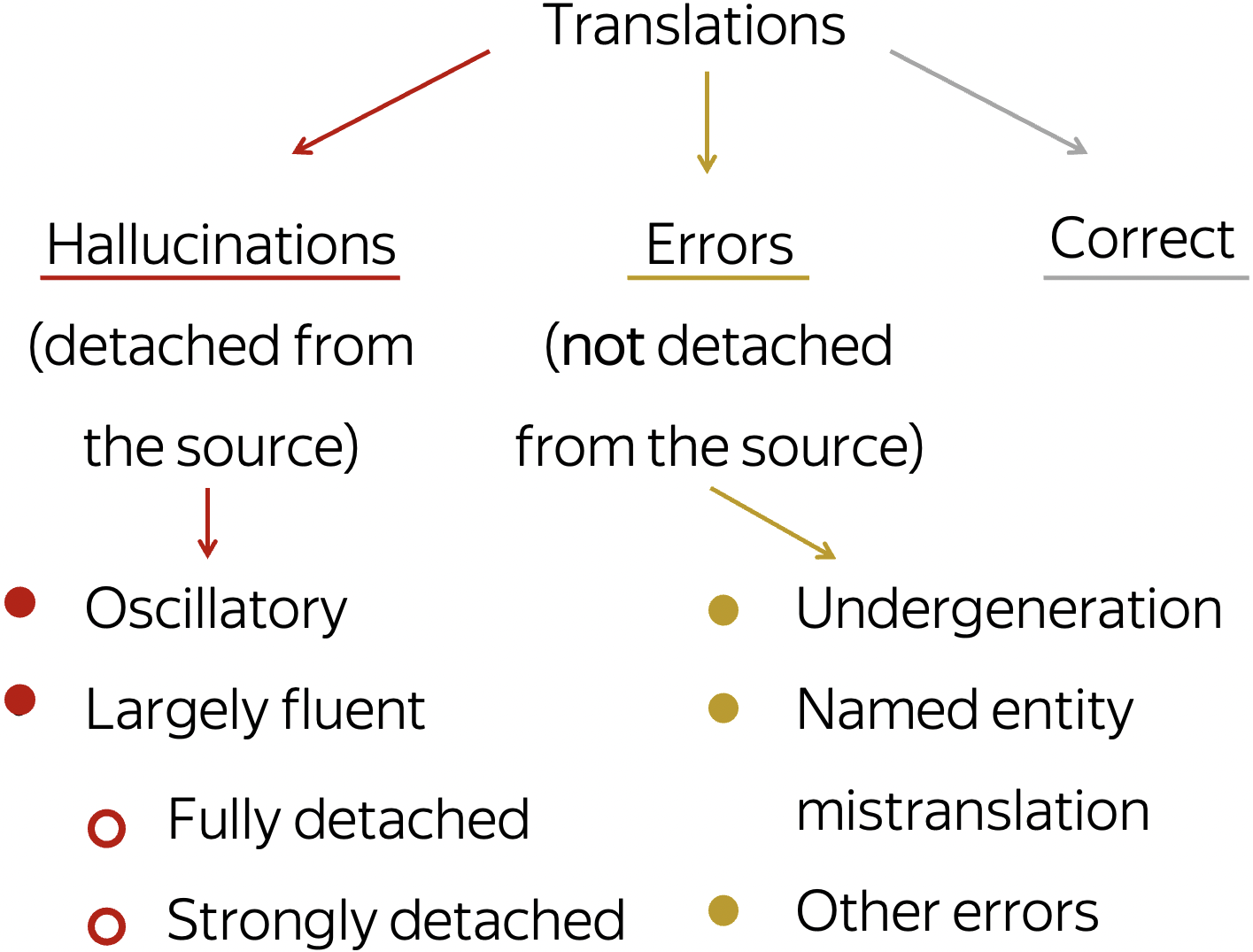}}
\caption{Taxonomy of translation types (based on the dataset by~\citet{guerreiro_hallucinations}).}
\label{fig:dataset}
\end{figure}

Hallucination dataset released by~\citet{guerreiro_hallucinations} contains fine-grained manual annotations of 3415 German-to-English translations generated by the model above. These translations are chosen from a set of 1.8M translations of held-out data as the ones that are likely to be pathological. The criteria used to flag the translations include 10~methods ranging from previously proposed heuristics~\cite{lee2019hallucinations,berard-etal-2019-naver,raunak-etal-2021-curious} to quality estimation models~\cite{rei-etal-2020-unbabels} and uncertainty detectors~\cite{fomicheva-etal-2020-multi,zerva-etal-2021-ist,guerreiro_hallucinations}. 

The taxonomy of translation pathologies in the dataset is shown in Figure~\ref{fig:dataset}. Here, hallucinations are defined as severe translation errors that are detached from the source. These can be either oscillatory (i.e. contain erroneous repetitions of words and phrases) or largely fluent. The latter is also split by severity of an error into fully detached (the whole content is not supported by the source) and strongly, but not fully, detached (significant proportion of output is not supported by the source).\footnote{\citet{guerreiro_hallucinations} mention that oscillatory hallucinations can also be either fully or strongly detached, but they do not divide this category into smaller groups because the overall number of such translations is rather small.} Other than hallucinations, the annotated data also contains translation errors that are deemed to be not detached from the source (see Figure~\ref{fig:dataset}).
Overall, 323 examples are judged to be hallucinations, 1044 as less severe translation errors and the rest as correct translations. 

Note that so far, there is no ``canonical'' hallucination taxonomy and previous work used various, mostly overlapping, definitions~\citep{lee2019hallucinations, raunak-etal-2021-curious, zhou-etal-2021-detecting, https://doi.org/10.48550/arxiv.2202.03629, salted_raunak2022,guerreiro_hallucinations}. We follow the taxonomy by~\citet{guerreiro_hallucinations} for two reasons. Firstly, for consistency with the dataset and the evaluation framework we use. Secondly, this taxonomy is rather general, considers some hallucination types overlooked in previous work (e.g. strongly detached hallucinations), and was shown to be reasonable: under these definitions,
properties of hallucinations differ from those of translation errors~\cite{guerreiro_hallucinations}.

\section{Hallucination Detection Methods}
\label{sect:criteria}

Generally, methods for handling hallucinations can be either \textit{internal}, i.e. using only information coming from the translation model itself, or \textit{external}, i.e. using auxiliary models. In addition to these, we also consider ``oracles'' relying on reference translation. Note that these cannot be used in preventive settings when references are not available; here we use them only for analysis.

\subsection{Reference-Based Oracles}

Following previous work~\cite{muller-sennrich-2021-understanding,guerreiro_hallucinations}, we use:
\begin{itemize}
    \item \textbf{chrF}: character $n$-gram F score of the translation with respect to reference. We use the \textsc{chrF++} version that also takes into account word unigrams and bigrams~\cite{popovic-2017-chrf};
    \item \textbf{COMET}: a neural quality estimation metric by~\citet{rei-etal-2020-comet} which was shown to be the state-of-the-art reference-based method~\citep{toshiptom}.
\end{itemize}

\subsection{Internal Measures}

\paragraph{Baseline: Seq-Logprob.} This is the standard length-normalized sequence log-probability. Compared to previously introduced methods specifically targeting hallucinations, this simple metric performs the best~\cite{guerreiro_hallucinations}.

\paragraph{We use ALTI: percentage of source contribution.}
As we already mentioned above, we hypothesize that the percentage of source impact on a generated translation might be a strong signal for identifying hallucinations. To evaluate this relative source contribution, we use recently introduced ALTI+~\cite{ferrandoALTI_plus}. At a high level, it decomposes each transformer block into a sum of functions of individual tokens and views an output representation as a summation of transformed input vectors. Then it evaluates contribution of these vectors to the resulting sum. Among other interesting observations, ALTI+ (as well as an earlier LRP-based method by~\citet{voita-etal-2021-analyzing}) was used to show that for artificially created hallucinations, 
source influence is much lower than for ``healthy'' correct translations. Our work is the first to test this intuition in a real setting where hallucinations are generated naturally.\footnote{Note that of the two methods that can evaluate relative source and target contributions we choose ALTI+ by~\citet{ferrandoALTI_plus} over LRP-based method by~\citet{voita-etal-2021-analyzing} because the latter is more computationally expensive. }

Formally, for a model and its generated translation, we compute the total source contribution as the sum of contributions of all source tokens. We do it for each target token individually and then average across target tokens.
The scores are computed by the same model that produced the translations~(Section~\ref{sect:model_guerreiro}).

\subsection{External models}

\paragraph{Baseline: COMET-QE.} For a reference-free model, we use the state-of-the-art \textsf{COMET-QE}~\citep{rei-etal-2020-unbabels} for its superior performance compared to other quality estimators~\citep{mathur-etal-2020-results, freitag-etal-2021-results, toshiptom}.

\paragraph{We use: sentence similarity.}
Overall, we consider three measures based on pretrained models that evaluate semantic similarity of two sentences:
\begin{itemize}
    \item \textbf{LASER}: cosine similarity of source and translation sentence embeddings from LASER2~\cite{heffernan_laser2}. LASER2 improves the encoder-decoder LASER~\cite{artetxe-schwenk-2019-massively} by replacing LSTM encoder with a Transformer and using teacher-student training;
    \item \textbf{LaBSE}: cosine similarity of source and translation sentence embeddings from LaBSE~\cite{feng-etal-2022-language}. LaBSE is a dual-encoder approach based on pretrained transformers and fine-tuned for translation ranking with an additive margin softmax loss; 
    \item \textbf{XNLI}: product of the entailment probabilities of source to translation and translation to source. We compute entailment scores with RoBERTa~\cite{conneau-etal-2020-unsupervised} fine-tuned on a combination of NLI data in 15 languages~\cite{conneau2018xnli}.\footnote{\url{https://huggingface.co/joeddav/xlm-roberta-large-xnli}}
\end{itemize}

\begin{figure*}[t]
\centering
{\includegraphics[scale=0.45]{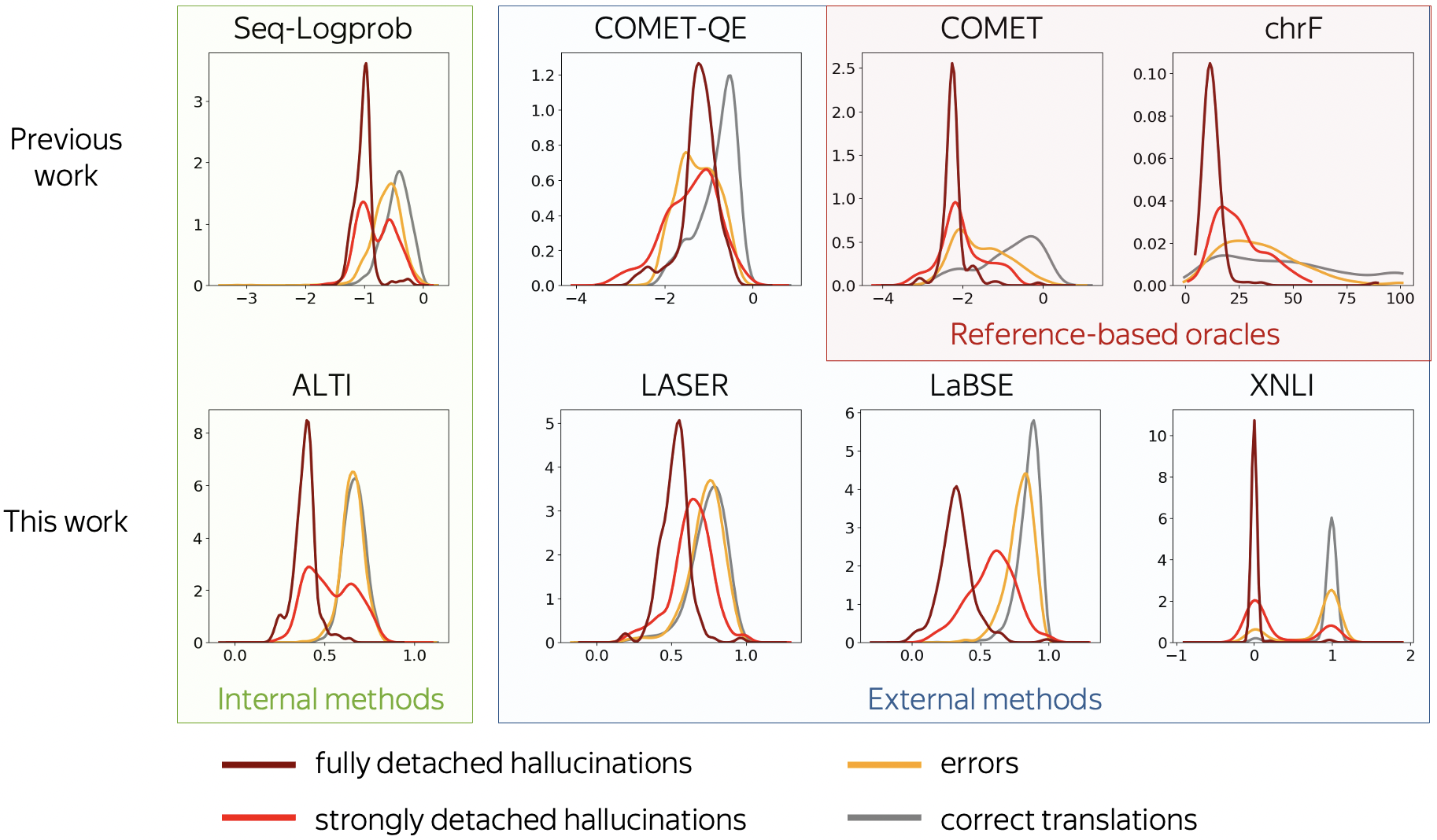}}
\caption{Kernel density estimation of the distribution of the detection criteria by translation pathology type. 
}
\label{fig:density}
\end{figure*}

\section{Detection Experiments}
\label{sect:detection_exps}

\subsection{Main results}
\label{sect:detection-main}

\begin{table}
\begin{tabular}{lcccc}
\toprule
 & \multicolumn{2}{c}{\bf All hall.} & \multicolumn{2}{c}{\bf Fully detached\!\!\!}\\
\!\!\!\textbf{Metric} & AUC  &  P@R90 & AUC & P@R90\!\!\! \\
        \toprule
        \!\!\!\textcolor{oracle}{ChrF} & \textcolor{gray}{75.4} & \textcolor{gray}{14.4} & \textcolor{gray}{89.6} & \textcolor{gray}{16.6} \\
        \!\!\!\textcolor{oracle}{COMET} & \textcolor{gray}{83.4} & \textcolor{gray}{19.2} & \textcolor{gray}{87.7} & \textcolor{gray}{12.6} \\
        \hline
        \!\!\!\textcolor{internal}{Seq-Logprob} & 83.0 & 13.9 & 93.5 & 31.0 \\
        
        \!\!\!\textcolor{internal}{ALTI} & \colorbox{internal!10}{84.9} & \colorbox{oracle!5}{12.5} & \colorbox{internal!60}{98.7} & \colorbox{internal!60}{67.4} \\
        \hline
        
        \!\!\!\textcolor{external}{COMET-QE} & \colorbox{oracle!40}{70.2} & \colorbox{internal!10}{14.2} & \colorbox{oracle!50}{66.1} & \colorbox{oracle!80}{\phantom{0}6.0} \\
        
        \!\!\!\textcolor{external}{LASER} & \colorbox{oracle!20}{79.4} & \colorbox{internal!10}{14.4} & \colorbox{oracle!10}{91.2} & \colorbox{oracle!30}{20.8} \\
        
        \!\!\!\textcolor{external}{LaBSE} & \colorbox{internal!50}{91.7} & \colorbox{internal!70}{25.9} & \colorbox{internal!50}{98.5} & \colorbox{internal!70}{70.3} \\
        
        \!\!\!\textcolor{external}{XNLI} & \colorbox{internal!40}{90.9} & \colorbox{internal!50}{24.1} & \colorbox{internal!60}{98.7} & \colorbox{internal!40}{60.4} \\
        \bottomrule
    \end{tabular}
    \caption{Hallucination detection quality. Methods: \textcolor{oracle}{oracle}, \textcolor{internal}{internal}, \textcolor{external}{external}. Changes in scores are highlighted compared to Seq-Logprob.}
    \label{tab:detection_table}
\end{table}

Overall results are shown in Table~\ref{tab:detection_table}. We report ROC AUC and precision at $90\%$ recall.\footnote{This is different from~\citet{guerreiro_hallucinations} who compare recall at thresholds cutting off a specific percentage of the dataset. 
Instead, we rely on two metrics: ROC AUC that does not rely on specific thresholds and PR@R90 that covers a specific percentage of the hallucinations (in this case, 90\%) and then reports the resulting precision.} We show metrics for all hallucinations and fully detached hallucinations separately
because the latter are the most disastrous translation mistakes
that are potentially the easiest to detect.

First, let us look at internal methods. We see that while ALTI performs comparably to Seq-Logprob for all hallucinations, for fully detached hallucinations it has twice better precision than Seq-Logprob. A possible root of this discrepancy is that ALTI averages the source contribution over all generated tokens. Therefore, it is more efficient for detecting translations where all or most of the generated tokens are hallucinated. Note also that for fully detached hallucinations, internal ALTI performs almost on par with the best external methods.

Among external methods, LaBSE and XNLI substantially outperform previous best detector: for both all and fully detached hallucinations, their precision at $90\%$ recall is roughly twice better than that of Seq-Logprob. While such a good performance might be expected for LaBSE that evaluates cross-lingual sentence similarity (in a way, this might be seen as a measure of translation quality), results for XNLI are rather surprising: to the best of our knowledge, models optimized for XNLI have not been used in the context of machine translation. This suggests that looking at broader class of models and training objectives might be beneficial.

Note also the large difference between LaBSE and LASER: while the former shows big improvements compared to Seq-Lobprob, the latter noticeably lags behind. This is not surprising when looking at training objectives of the underlying models. In LASER2, the cross-lingual part of the objective uses cosine similarity between sentence encodings. Differently, LaBSE is trained as a translation ranking task and thus encourages ordering of translations by severity of an error more explicitly.

To further understand differences between detectors, we look at the distributions of the detection scores in Section~\ref{sec:detection_density} and the detected pathology types in Section~\ref{sect:detection-types}.

\subsection{Analysing Distributions of the Scores}
\label{sec:detection_density}

For each of the methods, Figure~\ref{fig:density} shows distributions of the scores for fully detached hallucinations, strongly detached hallucinations, less severe errors and correct translations.

\paragraph{Internal methods: partial hallucinations are bimodal.} ALTI and Seq-Logprob show similar behavior: errors are distributed similarly to correct translations, and the scores for partial (strongly detached) hallucinations have bimodal distribution. 
At a high level, for the model, some partial hallucinations ``look'' more like full hallucinations, and some -- more like errors. This observation can motivate future work: it would be interesting to understand which types of these hallucinations behave one way or another. Specifically, whether it depends on detachment or on more simple patterns such as e.g. the proportion of hallucinated tokens.


\paragraph{COMETs: blind to error severity.} We see that COMET and COMET-QE scores do not provide separation between hallucinations and less severe errors. This agrees with previous work noting that 
since quality estimation models are mostly trained on data that
lacks negative examples,
COMETs may be inadequate at evaluating poor translations in general~\cite{takahashi-EtAl:2021:WMT, sudoh-etal-2021-translation} and hallucinations in particular~\cite{guerreiro_hallucinations}.  What is also expected, is that compared to reference-free \comet{}, the overlap between the scores for correct and incorrect translations is much lower for reference-based COMET. ChrF exhibits a behavior similar to COMET.

\paragraph{LaBSE: ranks hallucination severity best.} LaBSE is the only detector with a clear order between full, partial hallucinations, and translations without hallucinations. As we mentioned before, this is expected because only LaBSE is trained to rank translations. Interestingly, for LASER, modes for the three distributions are also ordered; unfortunately, the distributions themselves overlap significantly which makes LASER not a good hallucination detector. Both LaBSE and LASER ignore most of the non-hallucinated translation errors.

\paragraph{XNLI: no middle ground.} Finally, we see that XNLI distributions are very peaky and concentrated around 0 and 1. This is expected: XNLI's decision is always binary. While this provides good separation between fully detached hallucinations and correct translations, it is hard to estimate severity of an error.

\begin{figure}[t]
\centering
{\includegraphics[scale=0.28]{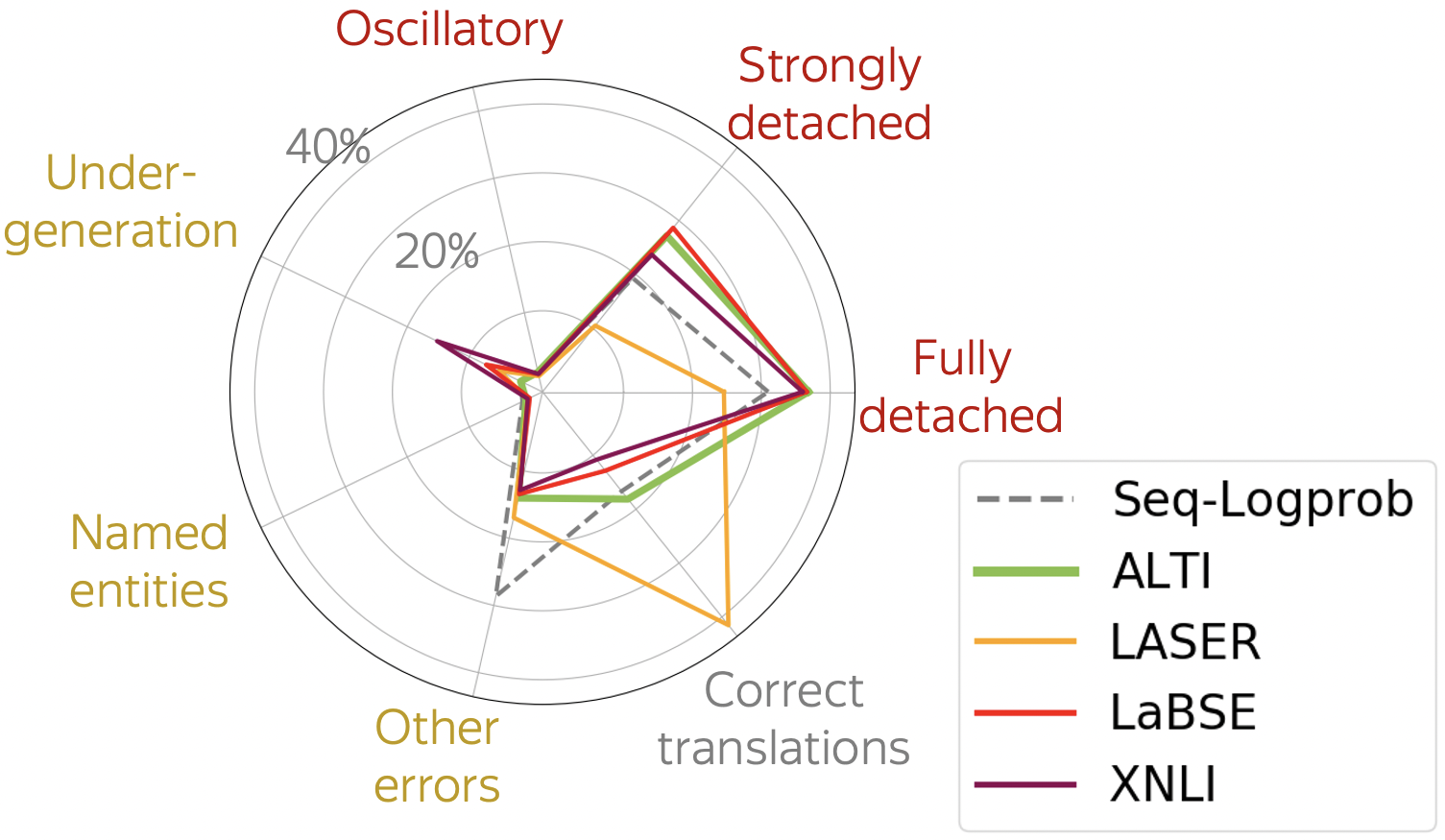}}
\caption{ Distribution of translation types when selecting the worst 10\% of the dataset according to each metric. In the original dataset, these types were annotated in a multilabel manner (e.g. the same translation could be annotated both as oscillatory hallucination and as a named entity error). To assign a single label to each translation, we choose the most severe pathology type (with severity increasing clockwise from ``Correct translations'' to ``Fully detached'').}
\label{fig:radar}
\end{figure}

\subsection{Detected Pathology Types}
\label{sect:detection-types}

\begin{figure}[t]
\centering
{\includegraphics[scale=0.29]{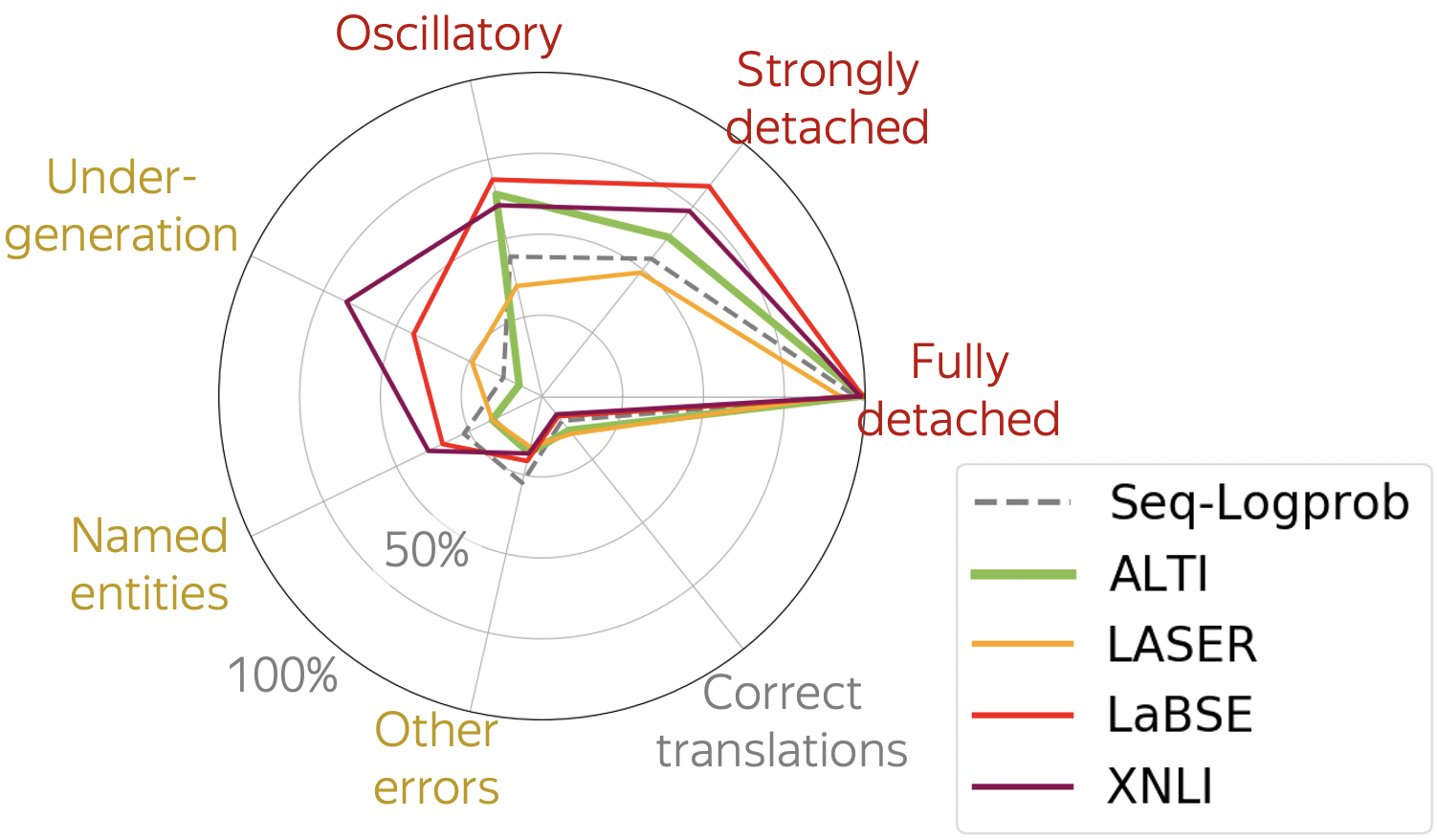}}
\caption{Recalls by translation types when selecting the worst 20\% of the dataset according to each metric. Here, the types are presented in a multilabel manner, i.e. one translation may contribute to multiple axes.}
\label{fig:radar_recall}
\end{figure}

Now we come to more fine-grained categories and look at detected pathology types. For each method, we say that a translation is ``detected'' if it is contained in a fraction (e.g. $10\%$) of the hallucination dataset corresponding to the lowest scores.\footnote{Note that we take such a large percentage because in the hallucination dataset we use, about $10\%$ of translations are hallucinations and about $30\%$ more are errors.} Then we look at
\begin{itemize}
    \item the distribution of pathology types contained among  detected examples (Figure~\ref{fig:radar});
    \item recall for different translation types with respect to the whole dataset (Figure~\ref{fig:radar_recall}).
\end{itemize}

\paragraph{The three best methods are similar.} Figure~\ref{fig:radar} shows that ALTI, LaBSE and XNLI select similar pathology types. For them, flagged examples consist mostly from fully detached and strongly detached hallucinations, along with other errors.

\paragraph{LASER is an outlier.} 
LASER behaves differently and instead of focusing on pathological translations, it flags correct translations more. This explains its poor performance on detection mentioned before.

\paragraph{XNLI flags undergenerations.} Figure~\ref{fig:radar_recall} shows that XNLI (and, to a lesser extent, LaBSE) flags a large proportion of undertranslations. This makes sense: these criteria are symmetric, and if we swap the source and the undergenerated translation, the longer source can be seen as a hallucination.

\paragraph{Fully detached are the easiest to detect.} As expected, fully detached hallucinations are the easiest to detect: all methods detect them entirely when taking $20\%$ of the hallucination dataset (Figure~\ref{fig:radar_recall}), and they are the most frequent pathology type among the examples flagged by the best performing methods (Figure~\ref{fig:radar}). Overall, our findings confirm the conclusions of ~\citet{guerreiro_hallucinations} that oscillatory and strongly detached hallucinations are more difficult to detect, and improvements with our methods mostly come from these types of hallucinations.



\begin{figure*}[t]
\centering
{\includegraphics[scale=0.45]{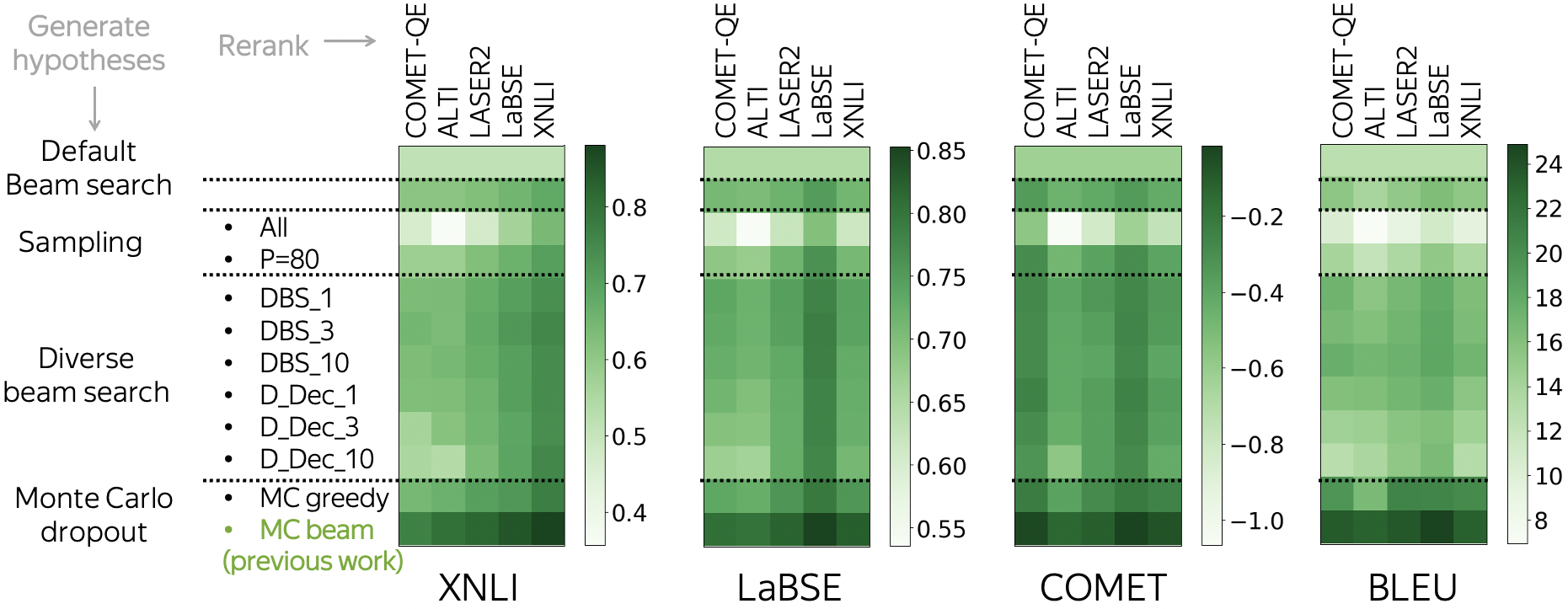}}
\caption{For all combinations of a generation strategy and a reranker, heatmaps show scores for the final translations.}
\label{fig:rerankers}
\end{figure*}

\section{Mitigating Hallucinations at Test Time}
\label{sect:mitigating}

Finally, let us come to the second part of the ``detect-then-rewrite'' pipeline: after flagging a translation as likely to be hallucinated, we want to generate several alternative translations and select one of them based on some criterion i.e., by reranking the hypotheses~\cite{guerreiro_hallucinations}. 
This general framework has two degrees of freedom: (i) generation of hypotheses, and (ii) reranking approach. 
We show that
\begin{itemize}
    \item for generating hypotheses, simply applying MC dropout (as done in~\citet{guerreiro_hallucinations}) outperforms more involved methods such as diverse beam search (Section~\ref{sec:generating_hypotheses});
    \item for reranking, we can match COMET-QE with internal ALTI and improve the hallucination rate by using LaBSE
    (Section~\ref{sec:reranking_approaches}).
\end{itemize}


\subsection{Evaluation methodology}

In this section, we explain the setup for the experiments with automatic evaluation in Sections~\ref{sec:generating_hypotheses} and~\ref{sec:reranking_approaches}. The setup for manual annotation is explained later in Section~\ref{sect:manual}.

\paragraph{Metrics.} In our experiments, we use several metrics. First, we use quality evaluation metrics commonly used by the community, i.e. COMET~\cite{rei-etal-2020-unbabels} and BLEU. Additionally, we use the two best metrics for hallucination detection: LaBSE and XNLI. We show some of the metrics in the main text and the rest in the appendix.


\paragraph{Data.} First, we analyze the impact of our method on translations of different quality levels. For this, we randomly sample 150 sentences from each of the following groups of the hallucination dataset (Section~\ref{sect:dataset}): fully detached hallucinations, strongly detached hallucinations, all other translation pathologies, and correct translations. We apply all versions of the hallucination mitigation algorithm to these 600 sentences.

Note that in a practical application, we would apply the mitigation techniques only to the translations labeled by a detection algorithm as potential hallucination. We simulate this later in Section~\ref{sect:manual} when performing manual annotation.


\subsection{Generation Strategies}
\label{sec:generating_hypotheses}

To generate alternative hypotheses, \citet{guerreiro_hallucinations} use Monte Carlo dropout~\cite{gal-mcdropout}. This means they leave standard beam search inference intact and achieve variability in translations via activating model dropout at inference. A natural question is whether using other generation strategies can give better results. For example, if we use e.g. beam search specifically designed to produce diverse translations, can we get better hypotheses? 

To test this, we use the following methods:
\begin{itemize}[noitemsep,nolistsep]
    \item \textsc{default}: standard decoding without reranking, i.e. beam search with size 5, where we pick only the top 1 candidate;

    \item \textsc{beam search}: beam search with size~$n$;
    
    \item sampling from the predicted distribution: 
    \begin{itemize}
        \item[$\circ$] \textsc{sampling}: from the whole distribution;
        \item[$\circ$] \textsc{sampling p=80}: from the top $p=80\%$ of the distribution, i.e. nucleus sampling~\cite{Holtzman2020The};
        \end{itemize}
        
    \item diverse beam search: 
    \begin{itemize}
        \item[$\circ$] \textsc{dbs\_n}: method by~\citet{vijayakumar2016diverse} with beam widths $s=1,\ 3,\ 10$;
        \item[$\circ$] \textsc{d\_dec\_r}: diverse decoding with diversity rates $r=1,\ 3,\ 10$~\cite{li2016simple};
        \end{itemize}
    
 \item Monte Carlo dropout: 
    \begin{itemize}
        \item[$\circ$] \textsc{mc greedy}: $n$ iterations of greedy search with dropout;
        \item[$\circ$] \textsc{mc beam}: the method used in \citet{guerreiro_hallucinations}, i.e. $n$ iterations of beam search with dropout, each with size $10$.
        \end{itemize}

\end{itemize}
Unless stated otherwise, $n=10$ in all experiments.

\subsubsection{The Impact of Generation Strategy}

The results are shown in Figure~\ref{fig:rerankers}. To disentangle the effect of generation strategy from the subsequent reranker performance, we show the results for all combinations. 
As rerankers, we considered COMET-QE used in~\citet{guerreiro_hallucinations} and the methods proposed in Section~\ref{sect:criteria}.

We see that the \textsc{mc beam} method clearly outperforms all the other. This is interesting for two reasons. First, MC dropout is easy to use: one has to apply standard inference with dropout on without other changes to the implementation. Next, differently from modifying decoding strategies, here variability in hypotheses comes from model predictive uncertainty~\cite{gal-mcdropout,zerva-etal-2021-ist,guerreiro_hallucinations}. This is one more evidence that understanding model inner characteristics can be beneficial in various settings.

Based on these results, in what follows we generate hypotheses with beam search with MC dropout.

\subsubsection{The Impact of Number of Hypotheses}

\begin{figure}[t]
\centering
{\includegraphics[scale=0.3]{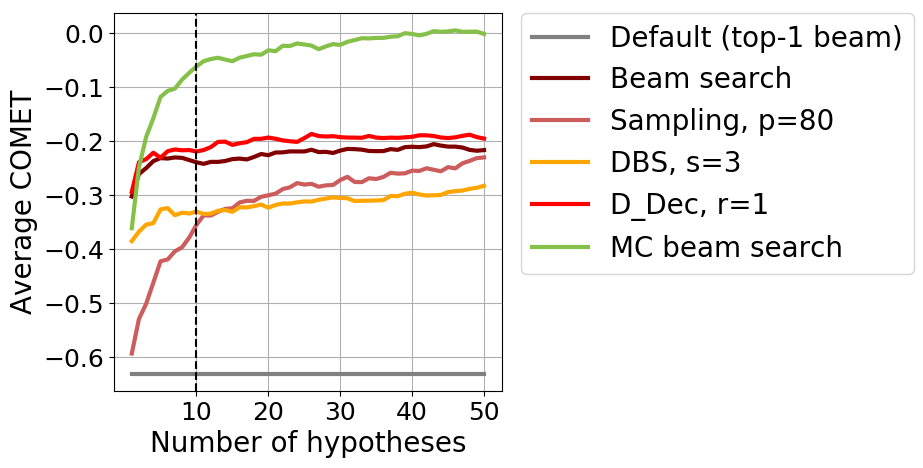}}
\caption{COMET scores for each generation method and number of hypotheses. For each group of generation strategies, we show the best representative.}
\label{fig:hyp_number_comet}
\end{figure}

We also check whether generating more than 10 hypotheses can improve the overall results. Figure~\ref{fig:hyp_number_comet} shows the final COMET scores depending on the number of hypotheses. We see that the scores increase with more hypotheses and do not saturate at 10. This implies that in cases when the quality of a translation is much more important than its computational cost, one can potentially improve the quality by generating more candidate hypotheses.

\subsection{Reranking Approaches}
\label{sec:reranking_approaches}

Apart from detecting hallucinations, the methods we propose can be applied as rerankers in the ``detect-than-rewrite'' pipeline. 


\subsubsection{Automatic Evaluation}

Figure~\ref{fig:rerankers} shows that, regardless of the generation method, LaBSE is the best reranker and it performs notably better than the strong COMET-QE baseline. Apart from the average results, Table~\ref{tab:comet_by_type} also shows COMET scores for each pathology type. 
We can see that reranking with any method is better then no reranking for all groups of original translations. Compared to the COMET-QE baseline, LABSE improves the scores for hallucinations and correct translations, but drops quality for other pathologies.

The only internal method ALTI performs better than COMET-QE
for fully detached hallucinations, but is inferior when looking at other translation types. It means that ALTI is very sensitive to the most severe pathology, but not capable to rank relatively good translations.


\begin{table}
\begin{tabular}{lcccc|c}
\toprule
 & \multicolumn{3}{c}{\bf Pathologies} & \!\!\!\!{\bf Cor.} & \!\!\!{\bf Avg.}\\
\textbf{Reranker} & \!\!\!F.  &  \!\!\!S. & \!\!\!O. &  \\
        \toprule
        No reranking & \!\!\!-1{.}23 & \!\!\!-0{.}97 & \!\!\!-0{.}59 & 0{.}27 & \!\!-0{.}63\\
        \!\!\!\textbf{Baseline} &  &  &  &  & \\
        COMET-QE & \!\!\!-0{.}21 & \!\!\!-0{.}13 & \!\!\!-0{.}14 & 0{.}35 & \!\!-0{.}03\\
        
        \!\!\!\textbf{Ours} &  &  &  &  & \\
        ALTI & \!\!\!-0{.}17 & \!\!\!-0{.}24 & \!\!\!-0{.}39 & 0{.}25 & \!\!-0{.}14\\
        LASER & \!\!\!-0{.}11 & \!\!\!-0{.}23 & \!\!\!-0{.}35 & 0{.}27 & \!\!-0{.}11\\
        LaBSE & \!\!\!-0{.}07 & \!\!\!-0{.}12 & \!\!\!-0{.}26 & 0{.}39 & \!\!-0{.}01\\
        XNLI & \!\!\!-0{.}12 & \!\!\!-0{.}18 & \!\!\!-0{.}28 & 0{.}30 & \!\!-0{.}07\\
        \bottomrule
    \end{tabular}
    \caption{Average COMET scores after reranking MC dropout hypotheses by various methods. Pathologies: fully detached hallucinations (F.), strongly detached hallucinations (S.), other pathologies (O.). In the appendix, we also show XNLI scores.}
    \label{tab:comet_by_type}
\end{table}

\subsubsection{Manual annotation}
\label{sect:manual}

\paragraph{Data.}
To confirm the results of automatic evaluation, we perform manual annotation. For each of the methods, we their translations of the same 200 source sentences. 
These sentences were randomly sampled from the hallucination dataset with the distribution of pathologies roughly mimicking outputs of the best detectors (Figure~\ref{fig:radar}). Overall, for 55$\%$ of the sentences their original translations are labeled as hallucinations, 25$\%$ as errors and 20$\%$ as correct translations.\footnote{We did not select these sentences with any metric, because it might affect the results of evaluating this metric as a reranker. We leave comparison of all combinations of detectors and rerankers to the future work.}

We compare the original translations and three reranking methods: the baseline COMET-QE used in~\citet{guerreiro_hallucinations}, the best overall reranker LaBSE, and the only internal method ALTI.

\paragraph{Annotation.} For each source sentence, the four translations were deduplicated and shuffled to avoid annotator bias. The resulting sentence pairs were given to 3 annotators who labeled them with three categories: Correct, Error, and Hallucination.  The labels were aggregated by majority vote; in case of ties (20 out of the 602 sentence pairs that were left after deduplication) we pessimistically assumed a hallucination. Further details on the annotation guidelines and inter-annotation agreement are reported in Appendix \ref{sec:appendix}. We evaluate the statistical significance of the pairwise differences in the proportions of correct and hallucinated translations using two-sided Student test for two related samples with 5$\%$ confidence level.

\begin{figure}[t]
\centering
{\includegraphics[scale=0.25]{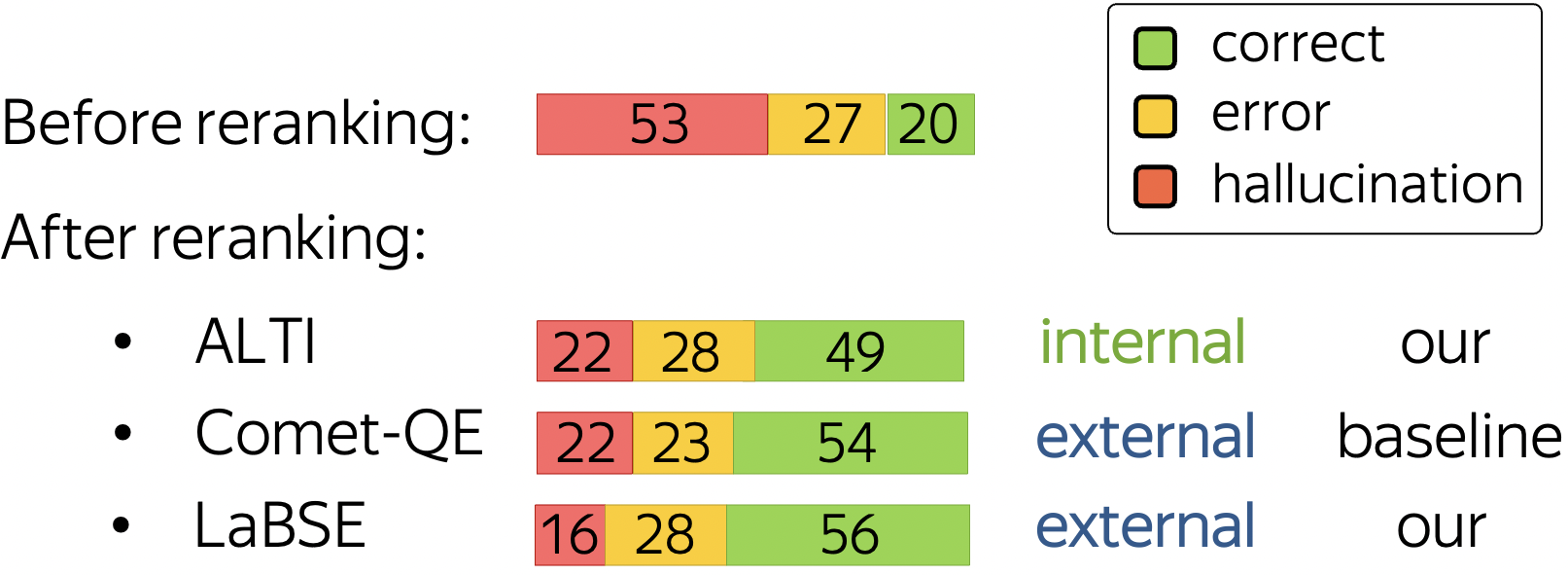}}
\caption{Human annotation results:
percentages of translation pathologies for different reranking methods. 
For hallucinations, all the differences are significant, except the one between ALTI vs COMET-QE. For correct translations, the difference between LaBSE and ALTI
is statistically significant. 
}
\label{fig:manual}
\end{figure}

\paragraph{Results.} Annotation results are shown in Figure~\ref{fig:manual}. All reranking methods reduce hallucinatory rate by a factor of 2.5-3. Interestingly, when looking at hallucinations,
internal ALTI performs on par with COMET-QE: the differences between these two methods are not statistically significant. COMET-QE, however, has less errors. This is expected: it was trained to distinguish correct translations from errors. Coming to LaBSE, we find that it produces slightly less hallucinations that other reranking methods, and more correct translations than ALTI; these differences are significant at 5$\%$ confidence level. Overall, by using sentence similarity from LaBSE, we improve both halucinations detection and alleviating hallucinations at test time. 

Note that since COMET-QE is the state-of-the-art quality estimator, it is a very strong baseline for the reranking stage where the goal is to find a better translation. The fact that we can match COMET-QE's hallucinatory rate reduction by analyzing model internal workings has value from different perspectives. For research, it can motivate future work on model understanding; for practitioners, it means that hallucination mitigation methods are not limited to language pairs where external models such as COMET-QE exist: model understanding might be enough.



\section{Conclusions}

We started this work by asking how far we can go at detecting and mitigating hallucinations if we use nothing but the translation model itself. Turns out, we can improve the results of the overall ``detect-then-rewrite'' pipeline by evaluating the percentage of source contribution to generated translation: translations with low source contributions are likely to be ``detached'' from the source, i.e. hallucinations. For detecting the most severe type of hallucinations, this method improves previous results twice; for mitigating hallucinations at test time, it matches the hallucination reduction rate of the previous method based on COMET-QE. We believe this can motivate future work on model understanding and help practitioners: using nothing but model inner working can allow mitigating hallucinations for language pairs where models such as COMET-QE are not available.

When allowing external models, we propose to expand the methods for handling hallucinations from models specialized for quality estimation to a broader set of objectives, e.g. sentence similarity from cross-lingual embeddings. Apart from showing that e.g. LaBSE improves previous results significantly, we also find that models so far overlooked in the context of machine translation (such as natural language inference) can be beneficial. We hope future work will build on this idea.

\bibliography{hal}
\bibliographystyle{acl_natbib}

\appendix

\section{Mitigating Hallucinations at Test Time}

Table~\ref{tab:xnli_by_type} shows XNLI scores after reranking MC dropout hypotheses by various methods.
Note that since here XNLI was used both to rerank and well as evaluate quality, in ths experiment XNLI can be viewed as an oracle.

\begin{table}
\begin{tabular}{lcccc|c}
\toprule
 & \multicolumn{3}{c}{\bf Pathologies} & {\bf Correct} & {\bf Avg.}\\
\textbf{Reranker} & F.  &  S. & O. &  \\
        \toprule
        No reranking & 2 & 30 & 80 & 93 & 51\\
        \!\!\!\textbf{Baseline} &  &  &  &  & \\
        COMET-QE & 59 & 69 & 85 & 93 & 77\\
        
        \!\!\!\textbf{Ours} &  &  &  &  & \\
        ALTI & 64 & 73 & 92 & 91 & 80\\
        LASER & 72 & 73 & 92 & 92 & 82\\
        LaBSE & 74 & 80 & 92 & 94 & 85\\
        XNLI (oracle) & 75 & 83 & 98 & 97 & 88\\
        \bottomrule
    \end{tabular}
    \caption{Average XNLI scores after reranking MC dropout hypotheses by various methods. Pathologies: fully detached hallucinations (F.), strongly detached hallucinations (S.), other pathologies (O.).}
    \label{tab:xnli_by_type}
\end{table}

\section{Manual Evaluation}
\label{sec:appendix}

In this appendix we describe the manual evaluation. First, we detail the simple guidelines that were presented to manual annotators. Second, we report the number of annotators and inter-annotation agreement. 
Third, we report the results of statisitical sigificance tests for comparing all the methods.


\paragraph{Guidelines} Annotators were provided with the guidelines shown in Table \ref{tab:ha}. For the reporting purposes, ``Partial hallucination'' was grouped together with ``Full hallucination'', and ``Undertranslation'' with ``Other''. 

\begin{table*}[!ht]
\small
\begin{tabular}{p{1.0\linewidth}}
\hline
\\
Each row of the data consists of the German source sentence, its reference English translation (it is not always accurate!), and 1 to 4 machine translation outputs. 
The machine translation outputs are presented in a random order, to exclude the possibility of bias toward any specific method. 

For each of the machine translations, you need to assign one of the following labels:

\begin{itemize}

    \item OK: An acceptable translation; it conveys the main meaning correctly and does not introduce extra meaning. Some details still may differ, and minor errors are acceptable.
    \item Partial hallucination: a part of the translation is unrelated to the source, or is related very indirectly, such as via a common topic.
    \item Full hallucination: most or all of the translation is unrelated to the source, or is related very indirectly.
    \item Undertranslation: there is no hallucinations, but a significant part of the source is not translated at all.
    \item Other: there are no hallucinations or undertranlsations, but there are other translation errors that make the translation unacceptable.

\end{itemize}
\\\hline
\end{tabular}
\caption{Human annotations Guidelines \label{tab:ha}}
\end{table*}


\paragraph{Inter-annotation agreement} We evaluated inter-annotation agreement by Fleiss' Kappa. For the three annotators and the three aggregated labels, it equals 0.57 on the 602 sentence pairs that were labeled (with the 5 original labels, it is 0.55). This may be interpreted as moderate agreement.

\paragraph{The differences} The Tables \ref{tab:manual_pvalues_ok} and \ref{tab:manual_pvalues_hall} compare proportions of correct and hallucinated translations for each of the manually evaluated methods. The P-values are computed with paired two-sided Student test (\texttt{scipy.stats.ttest\_rel}).


\begin{table}[!ht]
    \small
    \centering
    \begin{tabular}{llrrr}
        Method 1 & Method 2 & Rate 1 & Rate 2 & P-value \\
        \midrule
        LABSE & COMET-QE & 0.56 & 0.54 & 0.53 \\
        LABSE & ALTI & 0.56 & 0.49 & 0.02 \\
        LABSE & Default & 0.56 & 0.20 & 0.00 \\
        COMET-QE & ALTI & 0.54 & 0.49 & 0.12 \\
        COMET-QE & Default & 0.54 & 0.20 & 0.00 \\
        ALTI & Default & 0.49 & 0.20 & 0.00 \\
    \end{tabular}
    \caption{Comparison between manually annotated rates of correct translation.}
    \label{tab:manual_pvalues_ok}
\end{table}

\begin{table}[!ht]
    \small
    \centering
    \begin{tabular}{llrrr}
        Method 1 & Method 2 & Rate 1 & Rate 2 & P-value \\
        \midrule
        LABSE & COMET-QE & 0.16 & 0.22 & 0.01 \\
        LABSE & ALTI & 0.16 & 0.22 & 0.01 \\
        LABSE & Default & 0.16 & 0.53 & 0.00 \\
        COMET-QE & ALTI & 0.22 & 0.22 & 1.00 \\
        COMET-QE & Default & 0.22 & 0.53 & 0.00 \\
        ALTI & Default & 0.22 & 0.53 & 0.00 \\
    \end{tabular}
    \caption{Comparison between manually annotated rates of hallucinated translation.}
    \label{tab:manual_pvalues_hall}
\end{table}

\end{document}